\documentclass{Interspeech2024}

\interspeechcameraready

\title{Continual Learning Optimizations for Auto-regressive Decoder of Multilingual ASR systems}

\name[affiliation={1}]{Chin Yuen}{Kwok}
\name[affiliation={2}]{Jia Qi}{Yip}
\name[affiliation={2}]{Eng Siong}{Chng}

\address{
$^1$Digital Trust Centre, Nanyang Technological University, Singapore\\
  $^2$
College of Computing and Data Science, Nanyang Technological University, Singapore}
\email{kwok0062@e.ntu.edu.sg, jiaqi006@e.ntu.edu.sg, aseschng@ntu.edu.sg}


\begin{document}

\maketitle

\begin{abstract}
    
    Continual Learning (CL) involves fine-tuning pre-trained models with new data while maintaining the performance on the pre-trained data. This is particularly relevant for expanding multilingual ASR (MASR) capabilities. However, existing CL methods, mainly designed for computer vision and reinforcement learning tasks, often yield sub-optimal results when directly applied to MASR. We hypothesise that this is because CL of the auto-regressive decoder in the MASR model is difficult. To verify this, we propose four optimizations on the decoder. They include decoder-layer gradient surgery, freezing unused token embeddings, suppressing output of newly added tokens, and learning rate re-scaling. Our experiments on adapting Whisper to 10 unseen languages from the Common Voice dataset demonstrate that these optimizations reduce the Average Word Error Rate (AWER) of pretrained languages from 14.2\% to 12.4\% compared with Experience Replay, without compromising the AWER of new languages.
    
\end{abstract}

\section{Introduction}

Continual Learning (CL) involves adapting pre-trained models to new data without catastrophic forgetting (CF), where the model forgets about previously learnt knowledge and leads to degraded performance on the pre-trained data. This is particularly relevant for expanding multilingual ASR (MASR) capabilities. However, CL for MASR is challenging.

To adapt a model to new data without CF, numerous CL methods are developed with their designs focusing on different aspects: Replay-based methods like ER \cite{brignac2023improving} focus on managing the training data; Regularization-based methods like EWC \cite{kirkpatrick2017overcoming} and MAS \cite{aljundi2018memory} reduce model weights mismatch; Optimization-based methods like GEM \cite{lopez2017gradient} and A-GEM \cite{chaudhry2018efficient} constrain gradients used to optimize models;  LwF \cite{li2017learning} and DER\cite{buzzega2020dark} reduce model outputs mismatch through distillation \cite{truong2024emphasized}; Dynamic-architecture-based \cite{mallya2018piggyback,rusu2016progressive} methods  add new components to the model.

\begin{figure}[]
  \centering
  \includegraphics[width=\linewidth]{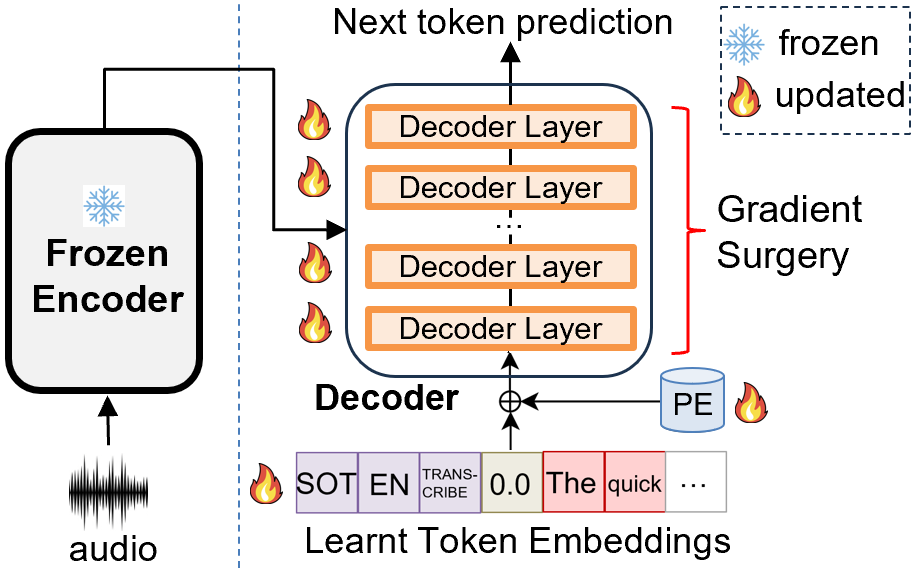}
  \caption{Overview of decoder-layer gradient surgery. For an MASR model that consists of an encoder (left) and decoder (right), we freeze the encoder and only adapt the decoder side. Also, we propose to apply gradient surgery only to the decoder layers, but not to the learnt token embeddings and learnt positional embeddings (PE).}
  \label{fig:decoder}
\end{figure}

As different CL methods have very different design focuses, previous works \cite{michieli2023online,cappellazzo2023investigation,vander2023rehearsal} have compared their effectiveness in the audio domain. Among these, some work specifically study CL methods to adapt MASR models to new languages. They include \cite{li2022massively}, which incrementally adds new languages and trains with all available data to mitigate forgetting, \cite{yu2023master}, which trains a mapping matrix for task-specific weight selection, and \cite{eeckt2022weight}, which uses weight averaging (AVG). Furthermore, the work by \cite{della2023cl} offers an extensive evaluation of baseline MASR CL methods, indicating potential areas for improvement. 

Notably, a significant research gap is that previous CL methods rarely focus their designs on the auto-regressive decoder model architecture of an MASR model. This is because most CL methods are primarily designed for computer vision (CV) and reinforcement learning (RL) tasks \cite{kirkpatrick2017overcoming,aljundi2018memory,li2017learning,brignac2023improving,buzzega2020dark,lopez2017gradient,chaudhry2018efficient} that make use of a different model architecture like the Convolutional Neural Network, and as a consequence, we show that directly applying these CL methods to MASR yield suboptimal results. We hypothesise that this is because the unique structure of the auto-regressive decoder presents unique CL issues.

They include the following: 1) The decoder uses token embeddings \cite{vaswani2017attention} to perform next token prediction, and the result is sensitive to the embeddings' norm and angular information \cite{demeter2020stolen}. Gradient-based CL methods which crudely modify these information by removing part of the gradients \cite{yu2020gradient} used to update the embeddings will cause sub-optimal results. 2) Most MASR models make use of the text embedding layer at the decoder which CV and RL models do not, although our experiments show that the embedding layer contributes significantly to CF. 3) The propagation of token output errors during auto-regressive decoding necessitates a meticulously designed strategy for the output of newly added tokens for new languages, ensuring that these errors do not adversely affect the model's performance. 4) The learning rate of the decoder needs to be meticulously tuned to prevent CF.

To address the four issues above, we propose four CL optimizations tailored for the auto-regressive decoder, which includes decoder-layer gradient surgery, freezing unused token embeddings, suppressing output of newly added tokens, and learning rate re-scaling to solve them respectively. Our contributions are four-fold. We show that 1) Averaged-GEM (A-GEM) \cite{chaudhry2018efficient}, which is a gradient-based CL method, can consistently outperform replay-based methods on MASR CL tasks if we only perform gradient surgery in the decoder layers and not on the token embeddings, 2) selectively freezing part of the text embedding layer at the decoder during training can reduce forgetting, 3) suppressing the output of newly added special tokens in the middle of the transcription can reduce error propagation, and 4) it is crucial to reduce the LR of the decoder rapidly during training to reduce forgetting.

\section{Methodology}
\begin{figure}[h!]
  \centering
  \includegraphics[width=\linewidth]{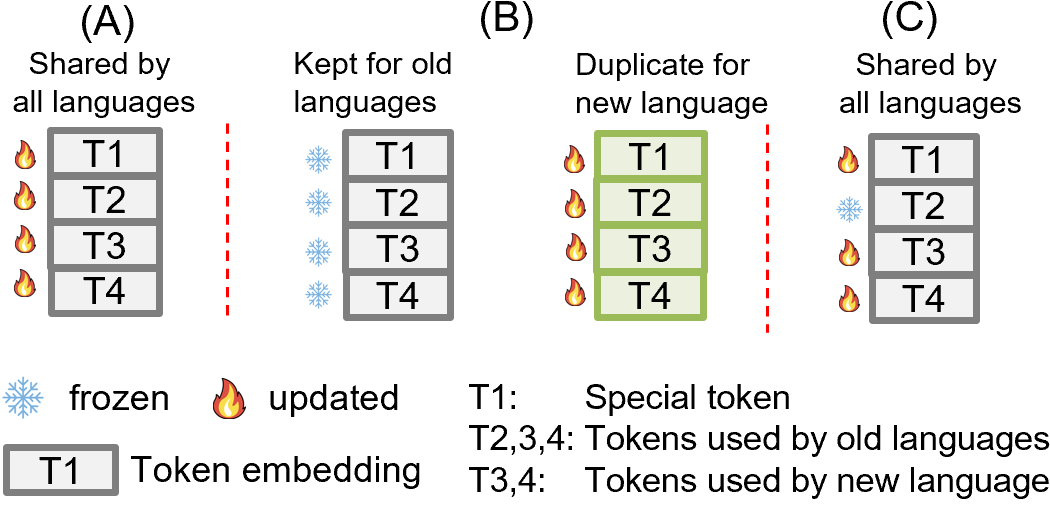}
  \caption{Strategy to adapt the token embeddings at the decoder to new tasks. All the embeddings are initialized from pre-trained weights. A) All the embeddings are shared, and updated for the old and new languages. B) A copy of the embeddings are adapted for the new langauge, and the original embeddings are kept for the old language. C) All the embeddings are shared, but only special tokens and tokens used by the new language are adapted.}
  \label{fig:embedding}
\end{figure}
\begin{figure}[t]
  \centering
  \includegraphics[width=\linewidth]{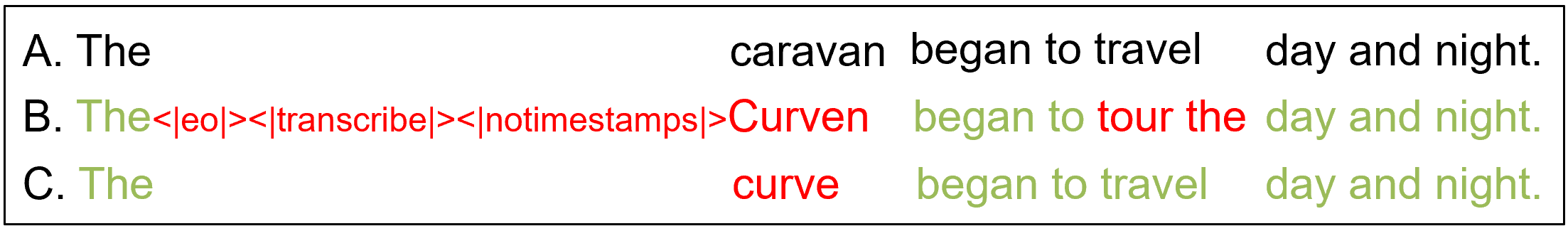}
  \caption{Example of Language ID Token Suppression (LID TS). A) Reference. B) Hypothesis before LID TS. C) Hypothesis after LID TS.}
  \label{fig:lidts}
\end{figure}

To learn unseen languages, we adapt the Whisper decoder and freeze the encoder \cite{della2023cl}. We train the model with A-GEM \cite{chaudhry2018efficient} with four optimizations for the decoder. First, as shown in Figure \ref{fig:decoder}, the gradient surgery from A-GEM is only applied to the decoder layers and not to the token embeddings at the decoder. Second, as shown in the right part of Figure \ref{fig:embedding}, we only update the embeddings of the special tokens and the tokens that appeared in the new languages.  Thirdly, as shown in Figure \ref{fig:lidts}, the output of special tokens at specific positions are suppressed. Lastly, we tune the Learning Rate Scheduler (LRS) to have a more repidly decreasing LR for the decoder during adaptation.

\subsection{Whisper Model for Language-agnostic MASR}

We use Whisper \cite{radford2023robust} as the base model for adaptation. This model supports MASR across 75 languages, including Language Identification (LID) functionality, making it language-agnostic \cite{datta2020language} as the user does not need to manually specify the language to transcribe. It utilizes a Transformer-based \cite{vaswani2017attention} attention encoder-decoder architecture with auto-regressive decoding. This model is selected because while comparative analysis reveals that Whisper's performance in Word Error Rate (WER) averaged across both seen and unseen languages is comparable to XLS-R \cite{conneau2020unsupervised} as reported by \cite{rouditchenko2023comparison}, \cite{della2023cl} indicate that Whisper outperforms WavLM \cite{chen2022wavlm} in overall MASR competency, and \cite{praveen2023language} demonstrates Whisper's superiority in LID over models based on wav2vec 2.0 \cite{babu2021xls} and WavLM. Also, the Whisper decoder is preferred for MASR applications over Large Language Model (LLM) based decoders \cite{tang2023salmonn}, particularly because Whisper decoder supports LID, a feature not available in the LLM-based decoder to the best of our knowledge.

\subsection{Experience Replay and A-GEM}

To mitigate CF, Experience Replay (ER) aims to remind a model of previously learnt tasks by incorporating a limited subset of data from previous tasks \cite{brignac2023improving} into the new task's dataset during adaptation. The goal is to minimize: \setlength\abovedisplayskip{10pt}
\begin{equation}\label{eqn:er}
\mathbb{E}_{(x',y')\sim{}D_B} [\ell(y', f_{\theta}(x'))] + \beta *\mathbb{E}_{(x'',y'')\sim{}\hat{D}_A} [\ell(y'', f_{\theta}(x''))]
\end{equation}
where $f_{\theta}$ is the model function parameterized by $\theta$, $\ell$ is the classification loss, $\hat{D}_{A}$ is a subset of the old task A dataset, $D_B$ is the new task B dataset, and $\beta$ is a hyper-parameter balancing the trade-off between the terms.

The approach is often combined with additional optimization constraints. Gradient Episodic Memory (GEM) \cite{lopez2017gradient} and Averaged-GEM (A-GEM) \cite{chaudhry2018efficient} use gradient surgery \cite{yu2020gradient} to constrain gradient update to not increase old task loss. Loss gradient vectors of old task $g_A$ and the gradient update on the new task $g_B$ is computed from $D_{\hat{A}}$ and $D_B$ respectively. To not increase old task loss while optimizing new task loss, the gradient $\hat{g}$ is proposed to minimize:
\begin{equation}\label{eqn:gem}
\frac{1}{2} ||g_B - \hat{g}||^{2}_{2}\quad \textrm{s.t.} \langle{}\hat{g}, g_A\rangle \ge{} 0 
\end{equation} 
Equation \ref{eqn:er} and \ref{eqn:gem} can be easily extended to settings with more than one old task by adding extra terms for each old task correspondingly.

\subsection{Gradient Surgery Removal}
Although the use of gradient surgery for CL has been proven effective in the CV field \cite{lopez2017gradient,chaudhry2018efficient}, its effect in the audio field is not better than other CL baseline methods like ER when an auto-regressive decoder is used \cite{vander2023rehearsal,della2023cl}. We hypothesise that this is because the decoder uses token embeddings to perform next token prediction, and the result is sensitive to the embeddings' norm and angular information \cite{demeter2020stolen}. Gradient-based CL methods which crudely modify these information by removing part of the gradients \cite{yu2020gradient} used to update the embeddings will cause sub-optimal results.

Therefore, to address the issue, we propose to remove A-GEM's gradient surgery from the token embeddings \cite{vaswani2017attention}. We find that this modified version of A-GEM can consistently outperform ER in our experiments.

\subsection{Partial Embedding Update}

Furthermore, we propose another optimization for the decoder, which is to selectively freeze part of the token embeddings to mitigate CF. This is because MASR models employ sub-word units for word representation \cite{sennrich2015neural}, and the sub-words can be shared across languages. If the sub-word token embeddings \cite{ethayarajh2019contextual} are adapted to new languages, it can inadvertently overwrite the semantic information of previously learned languages, leading to CF.

To address this issue, as shown in the right part of Figure \ref{fig:embedding}, we propose to update only the embeddings used by all the new languages and freeze the other embeddings.

\subsection{Language ID Token Suppression}

Moreover, another optimization is applied to suppress the decoder's output. This is necessary as to adapt Whisper to unseen languages, new language ID tokens are added to the text embedding layer, and the tokens may lead to more errors in the ASR output. Specifically, as shown in Figure \ref{fig:lidts}, we find that these ID tokens may be wrongly transcribed in the middle of the ASR output, and induce more errors in later transcriptions. Therefore, to address this, we propose suppressing the output of the tokens such that these tokens, primarily used for LID, will not interfere with the ASR results.

\subsection{Rapid LR Reduction During Adaptation}
\begin{figure}[h!]
  \centering
  \includegraphics[width=0.8\linewidth]{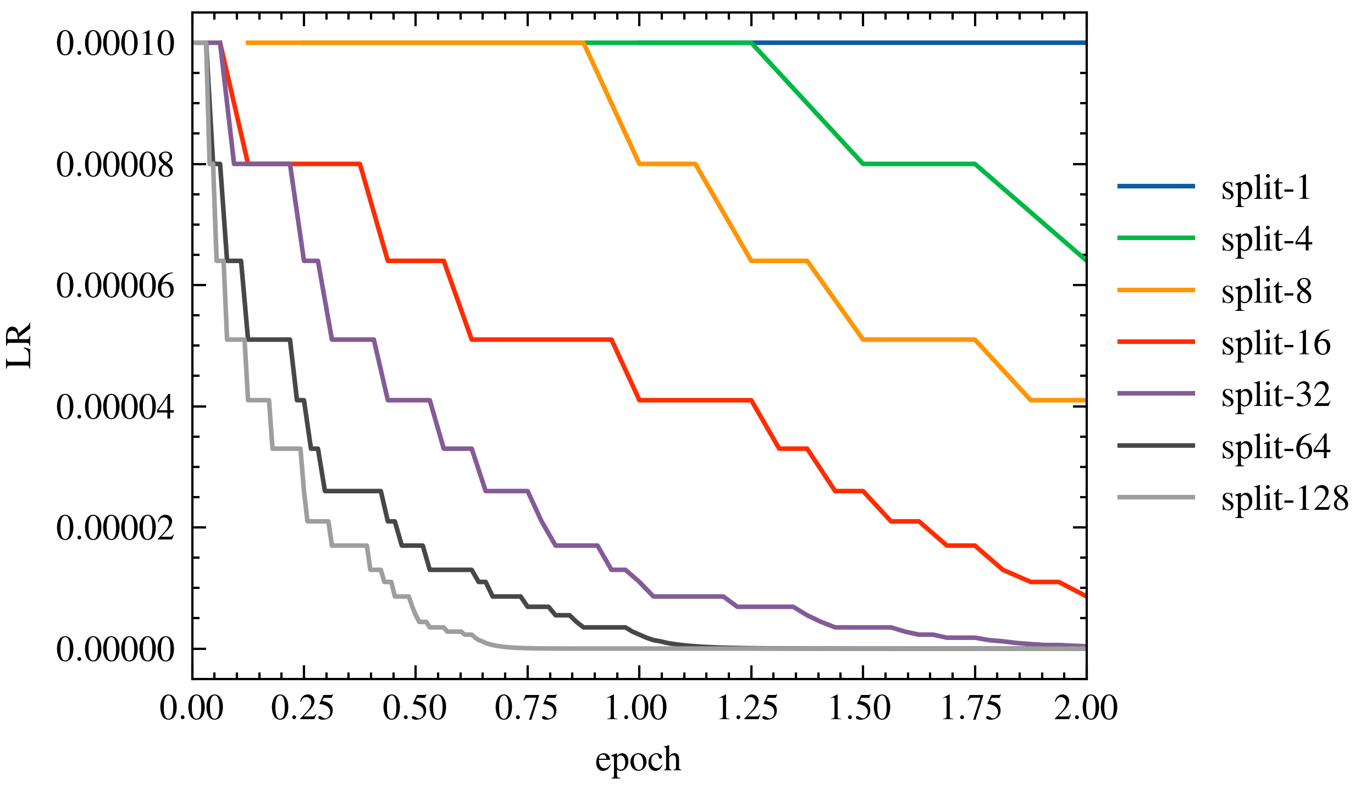}
  \caption{Change of LR during training as the validation interval changes. split-$n$ refers to validating every $1/n$ epoch.}
  \label{fig:val_interval}
\end{figure}

To further mitigate CF, we explore the effect of lowering the learning rate (LR) to constrain the deviation of the adapted model from its original state, thereby retaining more previously learnt knowledge. One method to modulate LR involves adjusting the Learning Rate Scheduler (LRS). Specifically, as our experiments use an LRS analogous to ReduceLROnPlateau\footnote{\url{https://pytorch.org/docs/stable/generated/torch.optim.lr\_scheduler.ReduceLROnPlateau.html}}, we shorten the validation interval such that the LR can be reduced more promptly by the LRS based on the more frequent validation feedback. Figure \ref{fig:val_interval} shows the effect of shortening the validation interval.

\section{Experiments}

\subsection{Dataset and Model Details}
We implement our methods based on the popular SpeechBrain \cite{ravanelli2021speechbrain} toolkit and CL-MASR \cite{della2023cl}.

Following previous works \cite{yu2023master,della2023cl}, we evaluate our method using a subset of the widely used large-scale CommonVoice dataset\footnote{\url{https://commonvoice.mozilla.org/en}} \cite{ardila2019common}. The subsets are extracted following \cite{della2023cl}. They consist of 10 languages pre-trained by Whisper and 10 new languages. Each language contains a 10 hours train set, a 1 hour validation set, and a 1 hour test set. whisper-small and whisper-large-v2 are adapted in two CL settings: 1) Adapt to ten pre-trained languages sequentially and test forgetting on ten new languages 2) Adapt to one pre-trained language and test forgetting on one new language. The former can take up to 6 days utilizing an NVIDIA A40 GPU.

We adapt the models for 2 epochs and set the train batch size to 4. Validation is performed every $1/32$ epoch for all methods. For ER, the replay data size is one hour for every old language. The AdamW optimizer \cite{loshchilov2017decoupled} is used with a variant\footnote{\url{https://speechbrain.readthedocs.io/en/latest/\_modules/speechbrain/nnet/schedulers.html\#NewBobScheduler}} of the ReduceLROnPlateau learning rate (LR) scheduler. For CL baselines and our methods, we sweep through the hyper-parameters to optimize average word error rate (AWER). A greedy decoding strategy is used for the experiments.

we refer to our methods as 1) ER-M, which is ER with our decoder optimizations except gradient surgery removal, and 2) A-GEM-M, which is A-GEM with all the decoder optimizations applied. All the experiments are tuned to have a rapid LR reduction during adaptation.

\subsection{Results and Discussion}
\definecolor{ao(english)}{rgb}{0.0, 0.5, 0.0}
\begin{table*}[]
\centering
\caption{WER of adapting Whisper-small to new language Interlingua (ia) and Esperanto (eo), and test forgeting on the pretrained language Germen (de) and English (en) respectively. Language-agnostic \cite{datta2020language} means the user does not need to specify the language to transcribe, and vice-versa for Language-aware. ``None" method refers to not performing adaptation.}
\begin{tabular}{ccccccccccc}
\toprule
\multirow{2}{1cm}{\textbf{Method}}                                            & \multicolumn{5}{c|}{\textbf{Language-aware WER (\%)}} & \multicolumn{5}{c}{\textbf{Language-agnostic WER (\%)}}\\
& \textbf{de}&  \textbf{en}  & \textbf{ia}& \textbf{eo}  & \multicolumn{1}{c|}{\textbf{avg}} & \textbf{de}& \textbf{en}  & \textbf{ia}& \textbf{eo}  & \textbf{avg}\\ \cmidrule{1-11} 
None                                                                                                                                                              &  14.00&             14.57  & n/a & n/a & \multicolumn{1}{c|}{n/a} & 14.00& 14.57  & n/a & n/a & n/a        \\
FT                                                                                                                                                                & 64.83&             68.99  & 12.31& 18.04  & \multicolumn{1}{c|}{$41.0_{\color{gray}{-00.0\%}}$} & 88.40& 90.55  & 12.31& 18.04  & $52.3_{\color{gray}{-00.0\%}}$      \\ \cmidrule{1-11} 
\textit{CL baselines}\\
AVG \cite{eeckt2022weight}                                                                                                                                                              & 17.94&             16.20  & 16.53& 38.96  & \multicolumn{1}{c|}{$22.4_{\color{ao(english)}{-45.4\%}}$} & 26.55& 18.37  & 16.53& 40.00  & $25.4_{\color{ao(english)}{-51.4\%}}$      \\
LwF \cite{li2017learning}                                                                                                                                                              & 16.81&             16.04  & 14.82& 22.20  & \multicolumn{1}{c|}{$17.5_{\color{ao(english)}{-57.3\%}}$} & 40.07& 74.96  & 14.82& 22.18  & $38.1_{\color{ao(english)}{-27.2\%}}$      \\
EWC \cite{kirkpatrick2017overcoming}                                                                                                                                                              & 23.78&             15.62  & 13.31& 19.21  & \multicolumn{1}{c|}{$18.0_{\color{ao(english)}{-56.1\%}}$ } & 41.12& 39.87  & 13.34& 19.21  & $28.4_{\color{ao(english)}{-45.7\%}}$       \\
MAS \cite{aljundi2018memory}                                                                                                                                                              & 18.29&             15.64  & \textbf{12.42}& 20.14  & \multicolumn{1}{c|}{$16.7_{\color{ao(english)}{-59.3\%}}$} & 21.97& 88.91  & \textbf{12.46}& 20.14  & $35.9_{\color{ao(english)}{-31.4\%}}$      \\
A-GEM \cite{chaudhry2018efficient}                                                                                                                                                            & 18.10&             18.15  & \textbf{12.42}& 18.74  & \multicolumn{1}{c|}{$16.9_{\color{ao(english)}{-58.9\%}}$} & 18.68& 19.09  & 12.68& 18.91  & $17.3_{\color{ao(english)}{-66.9\%}}$      \\
DER \cite{buzzega2020dark}                                                                                                                                                               & 16.31&             17.13  & 13.66& 19.54  & \multicolumn{1}{c|}{$16.7_{\color{ao(english)}{-59.3\%}}$} & 23.59& 56.95  & 13.66& 20.25  & $28.6_{\color{ao(english)}{-45.3\%}}$      \\ 
ER \cite{brignac2023improving}                                                                                                                                                                 & 16.65&             15.64  & 12.94& 20.34  & \multicolumn{1}{c|}{$16.4_{\color{ao(english)}{-60.0\%}}$} & 17.20& 18.26  & 13.14& 20.45  & $17.3_{\color{ao(english)}{-66.9\%}}$      \\ \cmidrule{1-11} 
\textit{our methods}\\
\begin{tabular}[c]{@{}l@{}}ER-M\end{tabular}                                          & \textbf{14.56}&             \textbf{14.96}  & 14.12& 19.88  & \multicolumn{1}{c|}{$15.9_{\color{ao(english)}{-61.2\%}}$} & \textbf{15.13}& 16.99  & 14.45& 20.06  & $16.7_{\color{ao(english)}{-68.1\%}}$      \\
\begin{tabular}[c]{@{}l@{}}A-GEM-M\end{tabular} 
& 15.37&             15.14  & 12.77& \textbf{18.39}  & \multicolumn{1}{c|}{$\bm{15.4}_{\color{ao(english)}{-62.4\%}}$} & 15.85& \textbf{15.96}  & 13.18& \textbf{18.56}  & $\bm{15.9}_{\color{ao(english)}{-69.6\%}}$ \\ \bottomrule
\end{tabular}
\label{tab:main_result}
\end{table*}

\begin{figure}[h!]
  \centering
  \includegraphics[width=\linewidth]{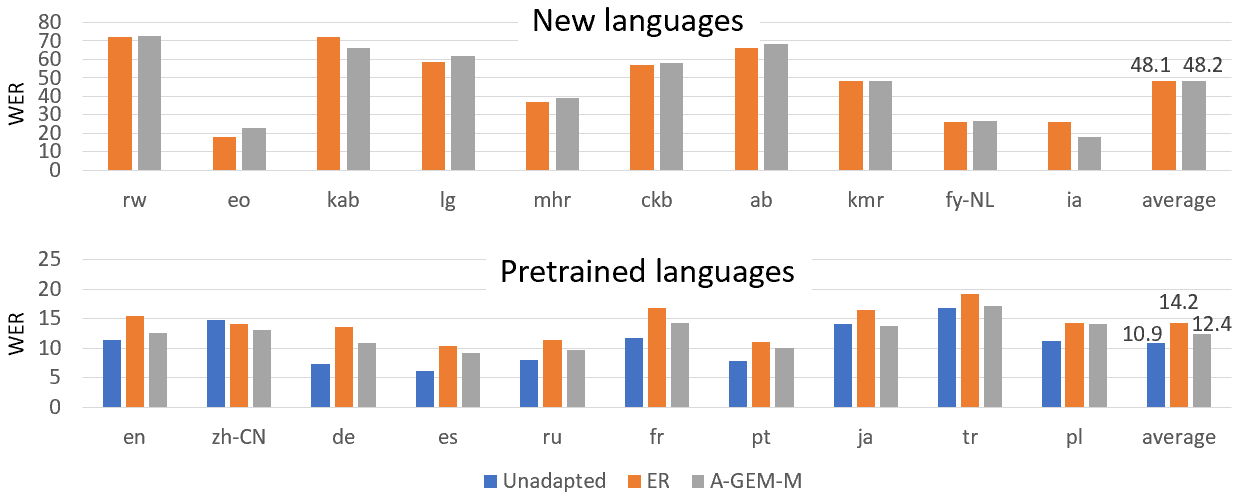}
  \caption{WER of transcribing 10 pre-trained languages and 10 new languages of varying difficulties. Results are obtained without manually specifying the language to transcribe. ``Unadapted" means the unadapted model.}
  \label{fig:wer}
\end{figure}

We show both the results of adapting to Whisper-small and Whisper-large-v2 models. Table \ref{tab:main_result} shows the results of adapting Whisper-small to two unseen languages separately. Vanilla full fine-tune (FT) can enable Esperanto and Interlingua ASR for Whisper, but leads to CF. All CL baselines reduce forgetting significantly and improve upon FT result by $27.2\%-66.9\%$ relatively. Among these, ER has the best overall performance, although EWC and MAS has slightly better partial results than ER for English and Esperanto in the language-aware setting. Weight averaging (AVG) performs less well than ER as we did not use a task-specific layer as in \cite{eeckt2022weight}.

Language-agnostic results are generally worse than language-aware results as the errors in LID may induce more error in transcription. Among the CL baselines, LwF, EWC, MAS and DER do not reduce forgetting as much as A-GEM and ER. We hypothesize that this is because the distillation of token-level logits for LwF and DER has no effect on the newly added language ID tokens. Similarly, the model weights' importance computed by EWC and MAS does not include the newly added language ID embedding weights, so its update is not regularized. As these methods fail to regularize the langauge ID tokens, they may cause more LID error, which in turn leads to more transcription errors. Additionally, we observe that when compared with language-aware results, the increase of WER in language-agnostic results mainly occurs in the old languages English and German. This suggests that the adapted Whisper model always bias its LID result of the input audios towards one of the newly adapted languages whenever it is uncertain.

Additionally, our methods outperform all CL baselines and improve the FT result by $61.2-69.6\%$. Forgetting is further reduced compared to all CL baselines while we can maintain similar WER for the newly adapted languages. Lastly in Figure \ref{fig:wer}, we plot the results of adapting whisper-large-v2 to the ten unseen language of various difficulties and test forgetting on the ten pretrained languages. Our method A-GEM-M outperforms the best CL baseline ER and reduce the Average Word Error Rate (AWER) of pretrained languages from 14.2\% to 12.4\% without compromising the AWER of new languages. We did not compare Dynamic-architecture-based methods \cite{rusu2016progressive,mallya2018piggyback} as they are not language-agnostic.

\subsection{Ablation study}

We further perform ablation studies and show the results in Table \ref{tab:ablation}. For the experiments, we adapt Whisper-small to the unseen language Esperanto (eo) and test forgetting on the pretrained language English (en). The results show that Gradient Surgery Removal reduces English WER from $18.15\%$ to $16.47\%$, showing that removing the gradient surgery in the text embedding layer reduces forgetting.

For Partial Embedding Update, it improves English WER from $15.46\%$ to $14.96\%$, showing that freezing tokens not used by the new language Esperanto reduces forgetting. We further try to naively train a completely separate text embedding layer at the decoder exclusively for Esperanto, and keep the original embedding layer for English. Results show that it further reduces WER slightly. However, the naive approach is not language-agnostic and requires more memory to store a separate embedding layer.

For Languae ID Token Suppression, it has no effect on English WER when no adaptation is performed. However, after adapting a model to the unseen language, it reduces English WER from $15.57\%$ to $14.48\%$ as it prevents the model from wrongly producing the newly added language ID token as the output, as shown in Figure \ref{fig:lidts}.

Lastly when we shorten the validation interval during training, it reduces English WER from $19.68\%$ to $15.64\%$ with not much effect on Esperanto WER. It shows that although tuning the validation interval is not crucial for the adapted language WER, it can reduce forgetting in the pre-trained languages by reducing learning rate more promptly whenever the validation loss is not decreasing, as shown in Figure \ref{fig:val_interval}.

\definecolor{ao(english)}{rgb}{0.0, 0.5, 0.0}
\mathchardef\mhyphen="2D 
\begin{table}[h!]
\centering
\caption{Ablation study of Languae ID Token Suppression, Partial Embedding Update, Gradient Surgery Removal and Shorten Validation Interval. ``None” method refers to not performing adaptation.}
\begin{tabular}{cccc}
\toprule
\multirow{2}{4cm}{\quad \quad \quad \quad \quad \textbf{Method}}& \multicolumn{3}{c}{\textbf{WER (\%)}}\\
& \multirow{1}{*}{\textbf{en}} & \multirow{1}{*}{\textbf{eo}} & \multirow{1}{*}{\textbf{avg}} \\ \cmidrule{1-4}
\multirow{1}{4cm}{\textit{Gradient Surgery Removal}} &&&\\
\multirow{1}{3.5cm}{A-GEM} & \multirow{1}{*}{18.15} & \multirow{1}{*}{18.74} & \multirow{1}{*}{18.4}\\
\multirow{1}{3.5cm}{\footnotesize{\quad + rm emb gradient surgery}} & \multirow{1}{*}{16.47} & \multirow{1}{*}{18.32} & \multirow{1}{*}{17.4} \\ \cmidrule{1-4}
\multirow{1}{4cm}{\textit{Partial Embedding Update}} &&&\\
\multirow{1}{3.5cm}{ER} & \multirow{1}{*}{19.68} & \multirow{1}{*}{20.53} & \multirow{1}{*}{20.1}\\
\multirow{1}{3.5cm}{\footnotesize{\quad + supress token}} & \multirow{1}{*}{15.46} & \multirow{1}{*}{20.34} & \multirow{1}{*}{17.9} \\
\multirow{1}{3cm}{\footnotesize{\quad + update partial emb}} & \multirow{1}{*}{14.96} & \multirow{1}{*}{19.88} & \multirow{1}{*}{17.5}\\
\multirow{1}{3cm}{\footnotesize{\quad + task-wise full emb}} & \multirow{1}{*}{14.48} & \multirow{1}{*}{20.32} & \multirow{1}{*}{17.4}\\ \cmidrule{1-4}
\multirow{1}{4cm}{\textit{Languae ID Token Suppression}} &&&\\
\multirow{1}{3.5cm}{None} & \multirow{1}{*}{14.57} & \multirow{1}{*}{n/a} & \multirow{1}{*}{n/a}  \\
\multirow{1}{3.5cm}{\footnotesize{\quad + supress token}} & \multirow{1}{*}{14.57} & \multirow{1}{*}{n/a} & \multirow{1}{*}{n/a} \\
\multirow{1}{3.5cm}{ER} & \multirow{1}{*}{19.68} & \multirow{1}{*}{20.53} & \multirow{1}{*}{20.1}\\
\multirow{1}{3.5cm}{\footnotesize{\quad + update partial emb}} & \multirow{1}{*}{15.57} & \multirow{1}{*}{20.42} & \multirow{1}{*}{18.0}  \\ 
\multirow{1}{3cm}{\footnotesize{\quad + supress token}} & \multirow{1}{*}{14.48} & \multirow{1}{*}{20.32} & \multirow{1}{*}{17.4}\\ \cmidrule{1-4}
\multirow{1}{4cm}{\textit{Shorten Validation Interval}} &&&\\
\multirow{1}{3.5cm}{ER\footnotesize{ + val. every epoch}} & \multirow{1}{*}{19.68} & \multirow{1}{*}{20.53} & \multirow{1}{*}{20.1}\\
\multirow{1}{3.5cm}{\footnotesize{\quad + val. every 1/32 epoch}}& \multirow{1}{*}{15.64} & \multirow{1}{*}{20.34} & \multirow{1}{*}{18.0}
\\ \bottomrule
\end{tabular}
\label{tab:ablation}
\end{table}

\section{Conclusion}


To conclude, this paper provides four optimizations to adapt multilingual auto-regressive decoder ASR models in a continuous learning framework to new languages and ablation study has shown the effectiveness of the proposed methods.

\section{Acknowledgement}

This research is supported by the National Research Foundation, Singapore and Infocomm Media Development Authority under its Trust Tech Funding Initiative. Any opinions, findings and conclusions or recommendations expressed in this material are those of the author(s) and do not reflect the views of National Research Foundation, Singapore and Infocomm Media Development Authority.

\bibliographystyle{IEEEtran}
\bibliography{mybib}

\end{document}